\documentclass{article}

 \PassOptionsToPackage{numbers, compress}{natbib}
\usepackage[final]{nips_2016}
\usepackage[utf8]{inputenc} 
\usepackage[T1]{fontenc}    
\usepackage{hyperref}       
\usepackage{url}            
\usepackage{booktabs}       
\usepackage{amsfonts}       
\usepackage{nicefrac}       
\usepackage{microtype}      
\usepackage{url}
\usepackage{graphicx}
\usepackage{setspace}
\usepackage{caption}
\usepackage{subcaption}

\title{LesionSeg: Semantic segmentation of skin lesions using Deep Convolutional Neural Network}

\author{
  Dhanesh Ramachandram \\
  School of Engineering\\
  University of Guelph\\
  Guelph, ON N1G 2W1, Canada\\
  \texttt{dramacha@uoguelph.ca} \\
  \And
  Terrance DeVries \\
  School of Engineering\\
  University of Guelph\\
  Guelph, ON N1G 2W1, Canada\\
  \texttt{terrance@uoguelph.ca} \\
}
\begin{document}
\maketitle

\section*{Executive Summary}

We present a method for skin lesion segmentation for the ISIC 2017 Skin Lesion Segmentation Challenge. Our approach is based on a Fully Convolutional Neural Network architecture which is trained end to end, from scratch, on a small dataset. Our semantic segmentation architecture utilizes several recent innovations in deep learning particularly in the combined use of (i) atrous convolutions to increase the effective field of view of the network's receptive field without increasing the number of parameters, (ii) network-in-network $1\times1$ convolution layers to increase network capacity  and (iii) state-of-art super-resolution upsampling of predictions using subpixel CNN layers. We achieved a IOU score of 0.642 on the validation set provided by the organizers.

\section{Background}

One of the fundamental and challenging tasks in digital image analysis is segmentation, which is the process of assigning pixel-wise labels to regions in an image that share some high-level semantics, hence the term ``semantic segmentation''.  In skin lesion segmentation, the goal is to assign pixel-wise labels to regions in dermoscopy images that represents skin lesions, such as melanoma, seborrhoeic keratosis or benign nevus. 

Skin lesion segmentation is challenging due to a variety of factors, such as variations in skin tone, uneven illumination, partial obstruction due to the presence of hair, low contrast between lesion and surrounding skin, and the presence of freckles or gauze in the image frame, which may be mistaken for lesions. A successful lesion segmentation technique should be robust enough to accommodate these variability.

Skin lesion segmentation is a widely researched topic in medical image analysis\cite{celebi2015state,korotkov2012computerized,celebi2009lesion}. Until recently, most skin lesion segmentation approaches were based on hand crafted algorithms\cite{zhou2011gradient,yuan2009narrow,schaefer2011colour}. Such approaches require carefully designed pre-processing and post-processing approaches such as hair removal, edge-preserving smoothing and morphological operations. The robustness of the such approaches can be somewhat limited however, as each new scenario may require custom tuning.

An alternative approach to manually crafting segmentation algorithms is to instead leverage machine learning techniques to learn a model capable of successfully dealing with the numerous factors of variation. Specifically applicable to this application are artificial neural networks, which have made an impressive resurgence in recent years. The active research area, now
known as deep learning,  is currently enjoying interest and support not only from
the academic community, but also from industry.  The advances
in hardware performance and low costs involved have made it viable to analyse
very large data sizes using very deep neural network architectures in a
reasonable amount of time. Deep learning-based techniques have resulted in
 state-of-the-art performance for many practical problems, especially
in areas involving high-dimensional unstructured data such as in
computer vision, speech and natural language processing. Medical imaging problems have also been given much attention by deep learning researchers\cite{wang2016perspective} and has seen tremendous success for the related skin lesion classification problem\cite{esteva2017}. The notion of being able to train in an end-to end manner, without requiring any manual feature engineering or complicated hand-crafted algorithms, is very attractive indeed.

\section{Semantic Segmentation}
The deep learning approach to image segmentation is known as semantic segmentation. In contrast to low-level image segmentation, which operate purely on local image characteristics such as colour, shape and texture, deep semantic segmentation algorithms are trained using thousands of examples to recognize and delineate regions in an image corresponding to some high-level semantics. A convolutional neural network can be adapted to perform perform semantic segmentation by replacing the top layer\footnote{typically the softmax layer} of a classification network into a convolutional layer. As FCNN use downsampling implemented via the max-pooling or strided convolutions to capture context, these architectures often employ a single or several progressive upsampling layers which are used to upscale lower resolution pixel-wise predictions to match the dimensions of the input image. Ground truth segmentation masks provide pixel-wise labels for the segmentation task, now cast as a pixel-wise classification problem.

The fully convolutional neural network architecture was first proposed by Long et al.\cite{long2015fully}. Subsequently, a number of similar architectures have been reported in the literature\cite{chen2014semantic,badrinarayanan2015segnet}.

\begin{figure}[ht]
\label{fig1}
\centering
\includegraphics[width=0.7\textwidth]{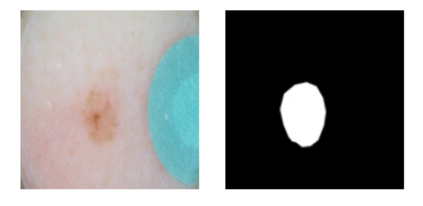}
\caption{Left: An example of a skin lesion image with a blue marker in the background. Right: The corresponding ground truth binary segmentation mask.}
\end{figure}

For the ISIC 2017 challenge, the skin lesion segmentation task is a binary segmentation task - the goal is to produce accurate segmentation of various skin lesions, benign and malignant, against  a variety of background which may consist of skin, colored markers, or dark-vignetted region produced by the dermoscope. Fig.\ref{fig1} shows an example lesion and its corresponding binary mask. The training dataset consists of 2000 dermoscopy images  and corresponding binary masks. The images consists of 3 types of skin lesions: nevus, seborrhoeic keratosis and melanoma; the latter lesion being malignant. The images are also of various dimensions. 

\section{Our Approach}

Our approach for lesion segmentation is primarily a fully convolutional neural network, trained from scratch, in an end-to-end manner. 

\subsection{Architecture}
\begin{table}[h]
\centering
\resizebox{0.85\textwidth}{!}{%
\begin{tabular}{@{}cccccr@{}}
\toprule
\textbf{Layer} & \textbf{Filter Size} & \textbf{Stride/Rate} & \textbf{Padding} & Non-linearity & \textbf{Type}      \\ \midrule
Conv1-1        & 5x5x64               & 2                    & Mirror           & ReLu          & 2D Convolution     \\
Conv1-2        & 3x3x96               & 1                    & Mirror           & ReLu          & 2D Convolution     \\
Conv1-3        & 1x1x96               & 1                    & Same             & ReLu          & 2D Convolution     \\
Conv2-1        & 3x3x128              & 2                    & Mirror           & ReLu          & 2D Convolution     \\
Conv2-2        & 3x3x256              & 1                    & Mirror           & ReLu          & 2D Convolution     \\
Conv2-3        & 1x1x256              & 1                    & Same             & ReLu          & 2D Convolution     \\
Conv3-1        & 3x3x256              & 2                    & Mirror           & ReLu          & Atrous Convolution \\
Conv3-2        & 3x3x256              & 2                    & Mirror           & ReLu         & Atrous Convolution \\
Conv3-3        & 3x3x128              & 2                    & Mirror           & None         & Atrous Convolution \\
Subpixel       & 3x3x32               & 1                    & Mirror           & None          & Subpixel CNN Layer \\ \bottomrule
\end{tabular}%
}
\caption{LesionSeg Architecture}
\label{arch}
\end{table}

The inputs to the network are images resized to $448\times448$. The first convolution layer uses a stride of 2 to downsample the image by a factor of 2.  This is followed by series of 2D convolution layers interspersed with $1\times1$ convolutions. We also apply batch-normalization for the convolutional layers and use ReLu activations. The $1\times1$ convolution layers add capacity to the network without increasing the number of parameters.The last two convolutions are dilated convolutional layers or \emph{atrous} convolutions\cite{chen2016deeplab} with a rate of 2. These two layers effectively enlarge the field of view of filters to incorporate larger context without increasing the number of parameters or the amount of computation.   Before upsampling is performed, the final subpixel convolution layer is added. This layer has number of filters equal to the upsampling factor and is applied with a stride of 1 and without any non-linearities. The upsampling layer is a subpixel convolutional layer introduced by \cite{shi2016real} which produced state-of-art super-resolution reconstruction accuracy superior to the commonly used bilinear upsampling as used by \cite{long2015fully}.

\subsection{Preprocessing}
The input images and the corresponding ground-truth masks are resized to 448 by 448 pixels. We perform data augmentation on-the-fly by randomly rotating both the image and its mask by 90-degree increments as well as flipping the images. In addition, we also perform per-image-standardization of the input image. 

\subsection{Training}

We trained the network using Adam\cite{kingma2014adam} optimization using a per-pixel cross-entropy loss function. During training, we performed randomly sampled images using a batch size of 32.  The network is trained until no improvement in the mean IOU is observed.

\subsection{Post-processing}
As the test and validation sets have images of different resolutions  up to $6688\times4439$ pixels, we resorted to upscaling the network output back to its original image dimensions using bicubic interpolation and then binarizing the upsampled output mask using a threshold of 128. We also apply morphological opening to eliminate small, spurious errors made by the semantic segmentation using a $3\times3$ disk-shaped kernel.

\section{Results and Discussion}

We achieved a IOU score of 0.642 for the validation set using our approach. Example segmentation outputs from our lesion segmentation architecture is shown in Fig.2.  

\begin{figure}[t!]
    \centering
    \begin{subfigure}[t]{0.45\textwidth}
        \centering
        \includegraphics[width=2in]{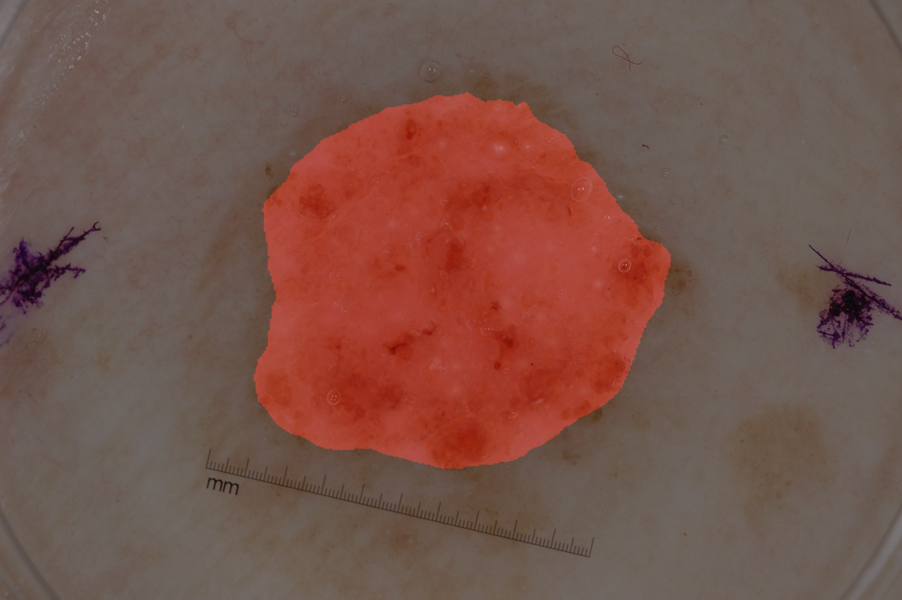}
        \caption{Sample 1}
    \end{subfigure}%
    ~ 
    \begin{subfigure}[t]{0.45\textwidth}
        \centering
        \includegraphics[width=2in]{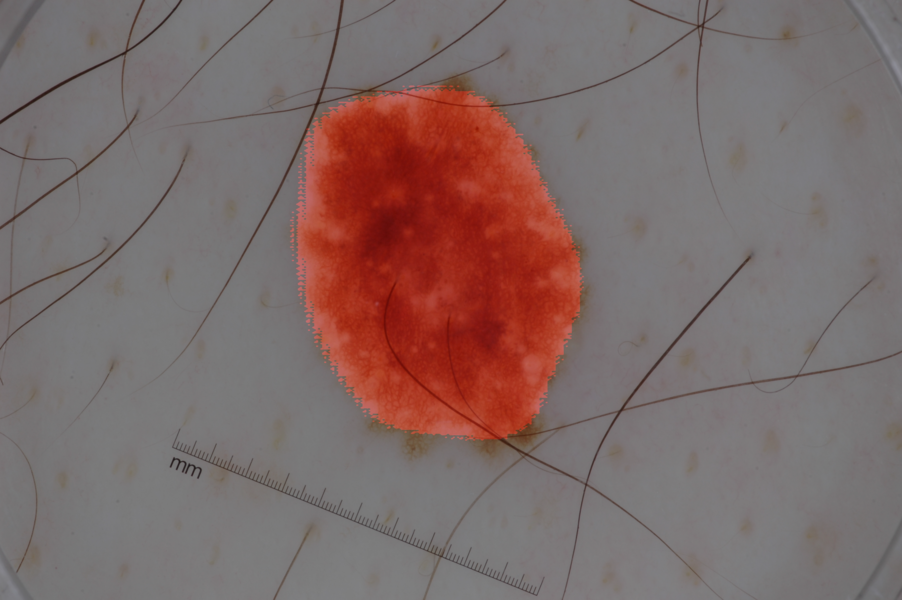}
        \caption{Sample 2}
    \end{subfigure}
    ~
     \begin{subfigure}[t]{0.45\textwidth}
        \centering
        \includegraphics[width=2in]{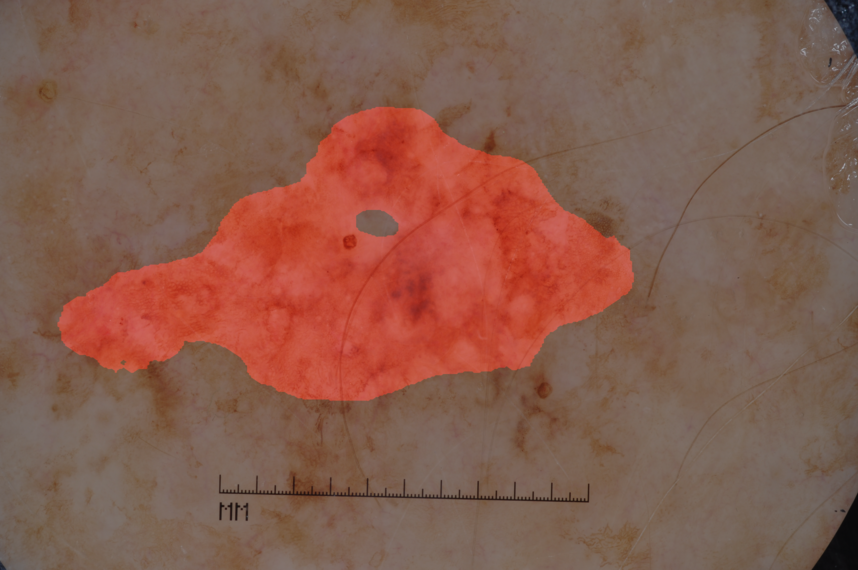}
        \caption{Sample 3}
    \end{subfigure}
     ~
     \begin{subfigure}[t]{0.45\textwidth}
        \centering
        \includegraphics[width=2in]{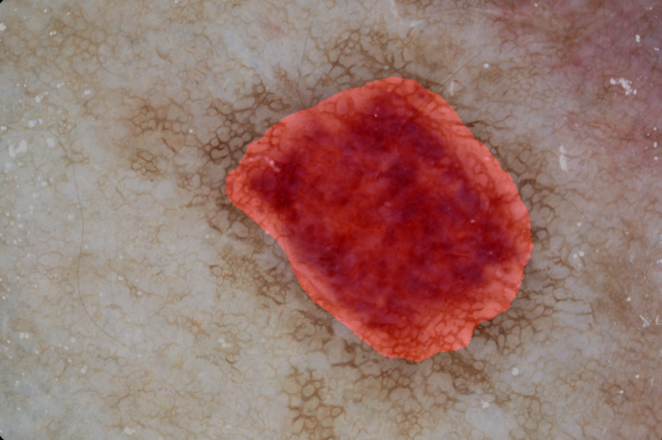}
        \caption{Sample 4}
    \end{subfigure}
    \caption{Examples of segmentation output using our approach.}
\end{figure}

\newpage
\bibliography{references}
\bibliographystyle{unsrt}

\end{document}